\documentclass{article}
\usepackage{cite}
\usepackage{amsmath,amssymb,amsfonts}
\usepackage{graphicx}
\usepackage{textcomp}
\usepackage{algorithm} 
\usepackage{algpseudocode} 
\usepackage{tablefootnote} 
\usepackage{caption}
\usepackage{subcaption}
\usepackage{multirow}
\usepackage{xcolor}
\usepackage{acronym}
\usepackage{hyperref}
\usepackage{authblk}

\acrodef{HDR}{High Dynamic Range}
\acrodef{LDR}{Low Dynamic Range}
\acrodef{VAM}{Visual Attention Module}
\acrodef{GMM}{Gaussian Mixture Model}
\acrodef{MAE}{Mean Absolute Error}

\title{High Dynamic Range Imaging via  Visual Attention Modules}

\author[1,2]{Ali Reza Omrani}
\author[1]{Davide Moroni}
\affil[1]{Institute of Information Science and Technologies (ISTI), National Research Council of Italy, Pisa, Italy}
\affil[2]{Department of Engineering, Università Campus Bio-Medico di Roma, Rome, Italy}
\affil[*]{\{ali.omrani, davide.moroni\}@isti.cnr.it}

\begin{document}

\maketitle

\section{Abstract}
Thanks to \ac{HDR} imaging methods, the scope of photography has seen profound changes recently. To be more specific, such methods try to reconstruct the lost luminosity of the real world caused by the limitation of regular cameras from the \ac{LDR} images. Additionally, although the State-Of-The-Art methods in this topic perform well, they mainly concentrate on combining different exposures and have less attention to extracting the informative parts of the images. Thus, this paper aims to introduce a new model capable of incorporating information from the most visible areas of each image extracted by a visual attention module which is a result of a segmentation strategy. In particular, the model, based on a deep learning architecture,  utilizes the extracted areas to produce the final \ac{HDR} image. The results demonstrate that our method outperformed most of the State-Of-The-Art algorithms.

\textbf{Keywords: }Deep Neural Network, High Dynamic Range imaging, Image Segmentation, Multi-exposure Image, Visual Attention Module

\section{Introduction}
\label{sec:Introduction}
In the scope of photography, the real world consists of an unlimited range of luminance. However, most devices are capable of capturing merely limited of that light. Therefore, the taken images are not desirable and consist of saturated regions, in which some parts of the images are too dark (underexposed) or overly bright (overexposed). These types of pictures are called \ac{LDR} images.

Thus, in order to cope with this problem, highly advanced cameras \cite{ 855857, 1467255, 4118491, 10.1145/2010324.1964936, Hajisharif2015, 7168378, 10.5555/3058909.3058929} can be used, which have special sensors to capture more light. However, such devices are mainly too expensive and overly heavy, which are not suitable for daily life, and instead, are mostly used in industries. 

A possible resolution for this drawback is developing software algorithms called \ac{HDR} imaging techniques. Moreover, \ac{HDR} images can be implemented by a single image \cite{10.1145/3130800.3130816, SHE2023105947, le2023singlehdr, 10.1145/3550277} or fusing a stack of images with different exposures, which are called single- and multi-exposure methods, respectively. In algorithms with a single \ac{LDR} image, an \ac{HDR} image can be produced from one image. However, the generated picture might not be as informative as the \ac{HDR} image produced by several \ac{LDR} images because the amount of detail in one single picture is limited compared to several images with different exposures. More precisely, \cite{10.1145/3130800.3130816} implemented an algorithm that only reconstructs the detail of bright saturated areas. However, the model is not only not capable of restoring the detail of dark regions but also does not perform well if the amount of bright saturation is too much. Thus, \cite{9116898} first combined several \ac{LDR} images and then fed the low-frequency response of the wavelet transform to the network to produce more detail in a shorter time.

Luckily, multi-exposure methods are more effective and informative compared to single-exposure techniques. Moreover, these methods perform well when the images are static \cite{8457442, 10.1145/3130800.3130834}, while when there are movements in the sequence of pictures, the ghosting problem emerges, which is almost solved in \cite{8803582, 10.1145/3072959.3073609, Wu_2018_ECCV, Yan_2019_CVPR, 8747329, 10.1007/978-3-030-58589-1_30}.

Deep learning has been a significant means of producing an \ac{HDR} image for the past decade. For instance, \cite{10.1145/3130800.3130816} produced an \ac{HDR} picture in the logarithmic domain with the help of a deep neural network. Additionally, \cite{8282319}, used a neural network to reconstruct the detail of an image with different exposure in each row in the irradiance domain. Moreover, unlike other multi-exposure methods, \cite{8457442, 10.1145/3130800.3130834} used a neural network to produce synthetic \ac{LDR} images with different exposures from a single image. Furthermore, \cite{10.1145/3072959.3073609} proposed to first align images with the help of the optical flow method, and then use a deep neural network to combine them. Therewith, \cite{8803582} instead of using optical flow for alignment, proposed to use two different neural networks first to align them and then combine the aligned images with the second neural network. Finally, \cite{8658831} used a neural network to learn the relative relation between the inputs and the Ground Truth using input images in different scales.

In this article, we would like to exploit image segmentation with the help of the Otsu method \cite{otsu1979threshold} in \ac{HDR} imaging to extract the most visible areas of the images and help the model produce pictures with more detail. Thus, to reach this point, \acp{VAM} will be proposed to obtain such regions.
Moreover, in this research, Spatial and Attention modules have been used from a State-Of-The-Art method, and a new architecture for the Reconstruction stage was designed and implemented, in which the visual attention and the reference image were used in the decoder part. Finally, although \acp{VAM} helped in producing pictures with more details and outperformed most of the State-Of-The-Art methods, the results still illustrated a slight amount of noise that was extracted from the input images.

In section \ref{sec:related}, the State-Of-The-Art in \ac{HDR} imaging and related image segmentation is presented. In section \ref{sec:method}, the proposed method is discussed in detail. Section \ref{sec:exp} demonstrates the experimental results and comparison with the State-Of-The-Art methods. Moreover, section \ref{sec:conclusions}  concludes this article with ideas for further works. Finally, the code will be available at the \href{https://github.com/AlirezaOmrani95/HDR-VAM}{\textbf{github page}}.

\section{Related work}
\label{sec:related}
In this section, we will discuss the State-Of-The-Art methods in the scope of \ac{HDR} imaging in the Multi-Exposure category (Section \ref{subsec:multi}) and survey unsupervised Image Segmentation methods for extracting regions (Section \ref{subsec:seg}).

\subsection{Multi-Exposure Methods}
\label{subsec:multi}

\cite{Li_2022_CVPR} proposed a two-stage algorithm, in which the first phase they extracted features from the input images, and merged them to produce the \ac{HDR} image in the latter one. Additionally, to cope with the appeared noise from the gamma correction operation on input images, i.e. the gamma-corrected Short-Exposure image becoming similar to Medium-Exposure, they used a U-net to extract noiseless features from it. Moreover, \cite{Deng2022} implemented a model in which images with lower scales were used to reduce the consuming sources. Additionally, a novel loss function was defined to focus more on the motion. Furthermore, \cite{MarinVega2022DRHDRAD} forwarded features with different scales to deformable and spatial attention blocks to align images in the feature space and also extract the features of the specific areas of the input images. Moreover, \cite{ye2022learning} proposed a model that at first estimated the optical flow from the two input images in different scales and then fused them to produce the final output. In \cite{xiao2022multiscale}, the   features are extracted from different scales and then are processed by sampling and aggregation modules  to align the pixels of the non-reference features. 

The work \cite{9857164} implemented a baseline that had lower computational resources and acceptable results compared to the other State-Of-The-Art models. They used a dual attention module, which includes both spatial and channel attention modules, to cope with misalignment and to better learn the details of the produced areas. In \cite{9857192}, the authors proposed a model that first extracts features  from input images by multi-scale encoding modules and then produces an \ac{HDR} image by progressively dilated U-shape blocks. 

\cite{dai2022waveletbased} demonstrated that the ghosting problem is mainly in short-frequency signals, and therefore, they proposed a wavelet-based model to merge images in the frequency domain and avoid any ghosting problems. \cite{prabhakar2022segmentation} implemented an algorithm that extracted dynamic areas of the images with the help of image segmentation and applied two neural networks separately on the static and dynamic scenes. Finally, they merged the information to produce an \ac{HDR} image without ghosting. In \cite{9856939} a model based on bidirectional motion estimation was proposed, in which, the amount of optical flow between \ac{LDR} images was estimated by motion estimation with cyclic cost volume and spatial attention maps, and eventually, an \ac{HDR} image was produced with the help of the extracted local and global features. \cite{Messikommer_2022_CVPR} implemented the first multi-bracket \ac{HDR} pipeline using event cameras, in which they merged the extracted features of images and the events to produce an \ac{HDR} image. \cite{10.1007/978-3-031-19800-7_20} proposed a transformer-based baseline, in which they used a context-aware vision transformer to extract local and global features to model the movement of objects and the diversity of intensity.

\subsection{Image Segmentation}
\label{subsec:seg}
Image segmentation is a crucial task in computer vision, which tries to partition images into segments to analyze the pictures more easily. Additionally, image segmentation not only can be used for object recognition, detection, and medical purposes but also can be applied for extracting regions of pictures with more details. In \cite{10.1007/978-3-540-30125-7_5} images were analyzed in HSV color space to segment pixels based on Intensity or Hue value. Moreover, two image segmentation methods were proposed based on luminance: histogram division \cite{8672126} and clustering based on \ac{GMM} of histogram \cite{kinoshita_kiya_2018}. Furthermore, \cite{9568865} calculated an optimal valley point based on the slope between the histogram value of each pixel and the neighboring points, and used the computed valley point to segment regions. The literature on the topic is endless, depending on applications and methodologies, from level set methods \cite{mitiche2010variational} to graph cut \cite{yi2012image} to recent deep learning-based frameworks \cite{minaee2021image}.

\section{Proposed Method}
\label{sec:method}
\subsection{Overview}

As cited in \cite{9594668}, it 
might be beneficial to first segment images based on exposure information to extract the best and more detailed regions from the  Over- and  Under-Exposure regions and exploit this knowledge in reconstructing an \ac{HDR} image. Following this idea, in this paper, a model is proposed in which, with the help of image segmentation, regions with more detail are segmented first in the preprocessing stage. Finally, they are fed to the model along with the input images to produce an \ac{HDR} image with the help of \acp{VAM}. 

Generally, the model can be divided into several sections. Firstly, the input images are fed into the feature extraction module, and afterward, the extracted features enter the attention and spatial alignment modules to cope with any possible misalignment. Moreover, the input images with their corresponding masks go to the \ac{VAM} simultaneously to extract the visible areas of the \ac{LDR} images. Next, the outputs of the three modules are fed to the Reconstruction stage to produce the initial \ac{HDR} image. Finally, the generated outcomes with the features of the reference image enter the refinement section to construct the final \ac{HDR} image.

\subsection{Preprocess}

In this article, the inputs are three \ac{LDR} images with different exposures, and the image with Medium-Exposure is considered the reference image. Moreover, before feeding the input images to the model, they are first mapped to the \ac{HDR} domain with the help of gamma correction. Finally, they are concatenated channel-wise with their corresponding \ac{LDR} images.
\begin{equation}\label{1st equation}
    \hat{I_i} =\frac{(I_i)^\gamma}{t_i} \qquad \textrm{for} \,\, i=1,2,3
\end{equation}
Where \(t_i\) is the exposure time of \(I_i\). \(\gamma\) is the gamma correction parameter, which was 2.24, and \(\hat{I_i}\) is the gamma-corrected image.

\subsubsection{Segmentation}

Most of the present algorithms in \ac{HDR} imaging focus more on the approach of image production, but not many pay attention to how to extract the most helpful features. Thus, in this research, the regions of the pictures with more details are segmented and extracted
as a preprocess and finally are fed to the proposed model along with the LDR images as the inputs. 

Different methods, such as the neural network and Otsu method were used for the image segmentation stage; however, the neural network resulted in overfitting. Thus, the Otsu method has been selected to segment the visible areas of the pictures. Therefore, the images are converted into the YUV color space to calculate a threshold based on the histograms of Short- and Long-Exposure images.
\begin{equation}\label{2nd equation}
    \mathrm{thresh}_i = G(Y_i) \qquad \textrm{for}\,\, i=1,3
\end{equation}
In which \(Y_i\) is the luminance channel of the LDR image, \(G()\) is the Otsu function, and \(\mathrm{thresh}_i\) is the threshold value of image \(i\).

In the Short-Exposure image, because most of the pixels are dark, and the objective is to extract the regions with visible pixels, the values equal to or more than the threshold are considered one, and the rest are zero for the Short-Exposure mask.
\begin{equation}\label{3rd equation}
    \begin{cases}
        1 & p\geq \mathrm{thresh}_1\\
        0 & p<  \mathrm{thresh}_1\\
    \end{cases}
\end{equation}

Where \(\mathrm{thresh}_1\) is the threshold value of the Short-Exposure image, and \(p\) is the pixel.

On the other hand, because most of the pixels in the Long-Exposure image are saturated, and the visible pixels have the lowest values, the values that are less than the threshold were considered one, and the rest as zero in the Long-Exposure mask.

\begin{equation}\label{4th equation}
    \begin{cases}
        0 & p\geq \mathrm{thresh}_3\\
        1 & p<\mathrm{thresh}_3\\
    \end{cases}
\end{equation}

By doing so, the masks of the areas with more detail are extracted and can help to produce an \ac{HDR} image.

\begin{figure}[h!]
    \centering
    \includegraphics[width=6cm]{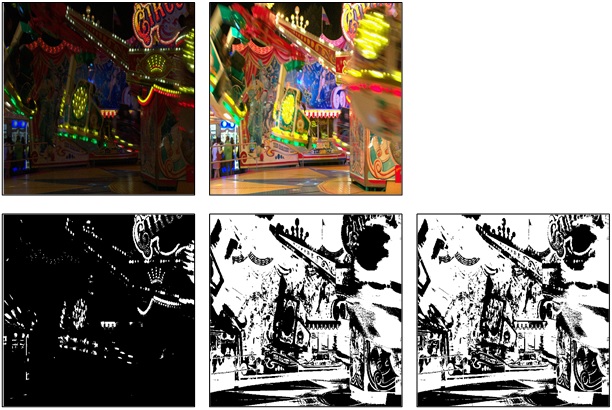}
    \caption{Produced masks of Short- and  Long-Exposure images.}
    \label{fig: Fig1}
\end{figure}

Generally, most of the pixels in Short- and Long-Exposure images are too dark or bright, respectively. Therefore, the location of the areas with surplus information is extracted and fed to the model. Doing so reduces the amount of calculation and helps in producing an \ac{HDR} image with more detail. Fig. \ref{fig: Fig1}  demonstrates the segmented and visible regions of both Short- and Long-Exposure pictures.

Moreover, during experiments, three input images with different exposures were used for image segmentation, in which, after obtaining the suitable areas of Short- and Long-Exposure images, the remaining regions were extracted from the Medium-Exposure image. However, the acquired areas of the Medium-Exposure were not sensible, as most of them were only a few pixels. Thus, two reasons exist for not using Medium-Exposure in the segmentation stage. First, it would be challenging to calculate a range for the visibility of the pixels. Second, Medium-Exposure is the reference image, and the picture will be used in the neural network. Therefore, it is not necessary to use segmentation for it.

\subsection{Proposed Method Structure}
\begin{figure*}[!ht]
   \centering
   \includegraphics[scale=0.29]{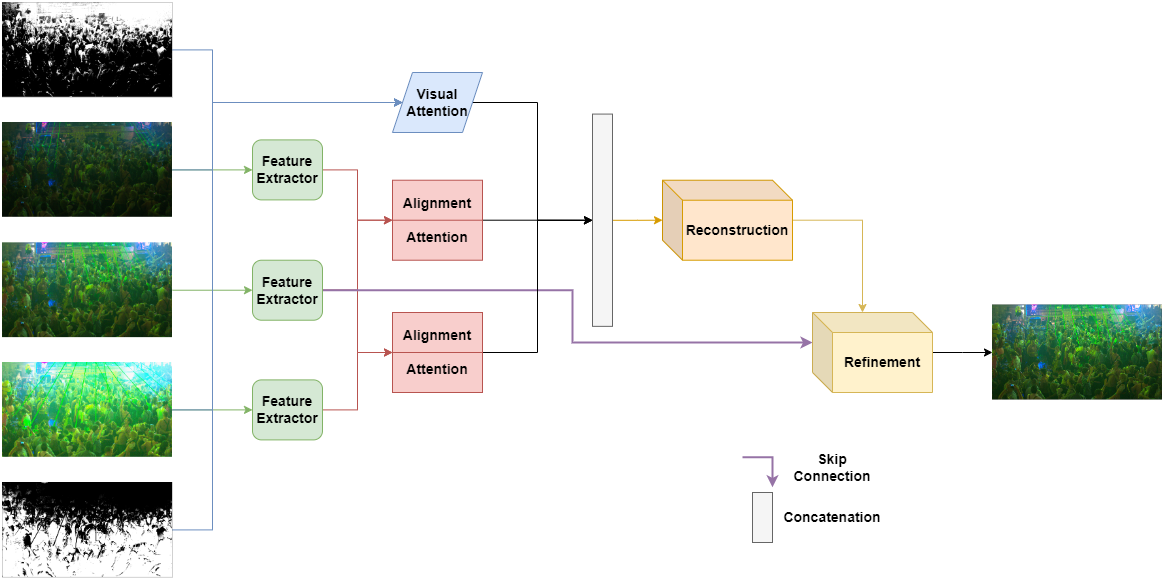}
   \caption{The total pipeline of the proposed.}
   \label{fig: Fig2}
\end{figure*}

As shown in Fig. \ref{fig: Fig2}, the proposed algorithm consists of six stages, which will be discussed separately and in detail.
\subsubsection{Feature Extraction}
\begin{figure}[!ht]
    \centering
    \includegraphics[scale=0.4]{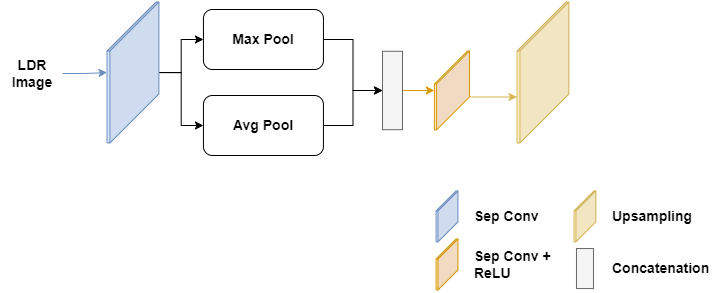}
    \caption{The structure of the Feature Extraction Block.}
    \label{fig: Fig3}
\end{figure}
Fig. \ref{fig: Fig3} illustrates the Feature Extraction block, in which a \(\mathrm{SepConv}\) is applied to the image to extract 32 feature maps. Afterward, a Max Pool and an Average Pool are used to not only smooth the features and focus on the details but also pay more attention to the edges. Next, the outputs of Poolings are concatenated, and another \(\mathrm{Sep\mathrm{Conv+ReLU}}\) is used to reduce the number of channels to 32. Finally, the extracted features are Upsampled to make them the same size as the input image. The feature extraction can be written as follows:

\begin{equation}\label{5th equation}
    C_i = \mathrm{concat}\left(M(\mathrm{SepConv}(I_i)),A(\mathrm{SepConv}(I_i))\right) 
\end{equation}
\begin{equation}\label{6th equation}
    F_i = \mathrm{Upsample}(\mathrm{ReLU}(\mathrm{SepConv}(C_i)))
\end{equation}

for $i=1,2,3$, where \(A()\) and \(M()\) functions are Max Pooling and Average Pooling, respectively, and \(C_i\) is the output of Concatenation. Finally, \(F_i\) is the output of the Feature Extraction Block.

\subsubsection{Visual Attention Module}

\begin{figure}[!ht]
    \centering
    \includegraphics[scale=0.4]{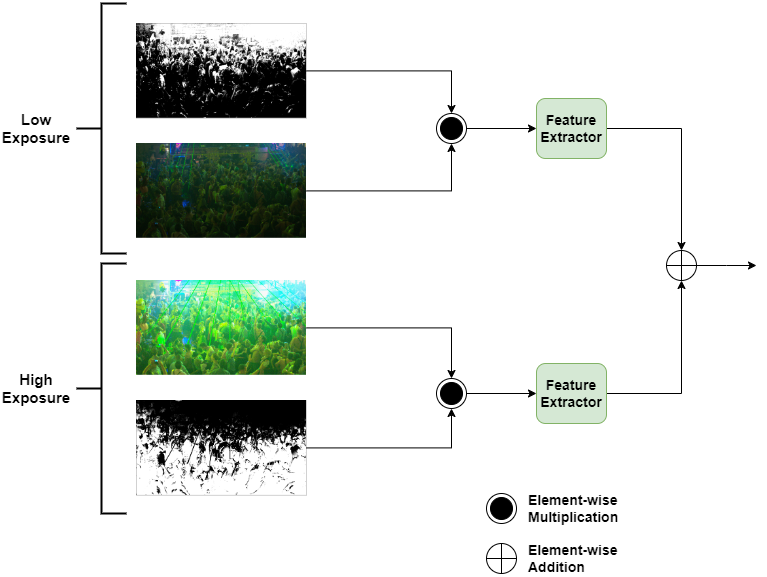}
    \caption{The structure of the Visual Attention Module (VAM).}
    \label{fig: Fig4}
\end{figure}

As it was mentioned, in this article, Image Segmentation is used to help the model to produce a better image. Therefore, as shown in Fig. \ref{fig: Fig4}, the input images are multiplied element-wise by their corresponding masks first. By doing so, the regions with more details are kept, and those that are overly dark or too bright will be removed. Next, they are fed to the Feature Extractor to extract Features. Finally, they are added together element-wisely. The \ac{VAM} can be formally defined as follows:
\begin{equation}\label{7th equation}
    \mathrm{features}_L=F(\mathrm{multiply}(\mathrm{mask}_L,I_L ))
\end{equation}
\begin{equation}\label{8th equation}
    \mathrm{features}_H=F(\mathrm{multiply}(\mathrm{mask}_H,I_H ))
\end{equation}
\begin{equation}\label{9th equation}
    V=\mathrm{add}(\mathrm{features}_L,\mathrm{features}_H )
\end{equation}

Where \(F\) is a feature extractor function, and \(V\) is the output feature of the VAM.

\subsubsection{Spatial Alignment Module}

\begin{figure}[!ht]
    \centering
    \includegraphics[scale=0.4]{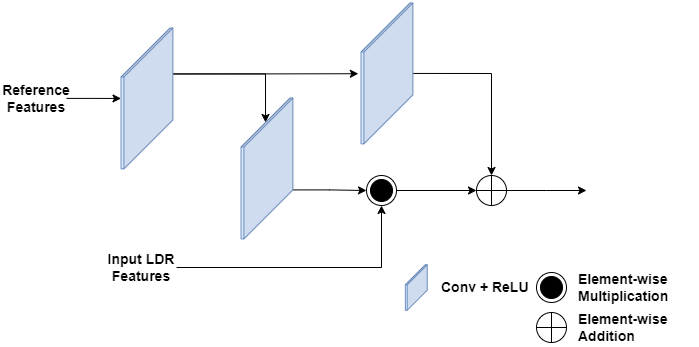}
    \caption{The structure of the Spatial Alignment Module.}
    \label{fig: Fig5}
\end{figure}

Because the input \ac{LDR} images are not aligned, the extracted features from the \ac{LDR} images without the gamma correction images are fed to an \emph{ad hoc} module for  aligning them. To this end, we used the same Feaure-alignment Module used in \cite{9857192}. As can be seen in Fig. \ref{fig: Fig5}, first a \(\mathrm{Conv+ReLU}\) is applied to the Reference Features, which can be called as \(\mathrm{Ref}_1\). Next, a \(\mathrm{Conv+ReLU}\) is applied to \(\mathrm{Ref}_1\) and is multiplied element-wisely by the input \ac{LDR} features, which can be called \(M_i\) (for $i=1,3$). Finally, another \(\mathrm{Conv+ReLU}\) is applied to the \(\mathrm{Ref}_1\) and is added element-wisely with \(M_i\). Formally, the operation in the module can be written as follows:

\begin{equation}\label{10th equation}
    \mathrm{Ref}_1 = \mathrm{ReLU}(\mathrm{Conv}(\mathrm{ref\, features}))
    \end{equation}
\begin{equation}\label{11th equation}
    M_i = \mathrm{multiply}(\mathrm{ReLU}(\mathrm{Conv}(\mathrm{Ref}_1)),\,\mathrm{inp\, features}_i)
\end{equation}
\begin{equation}\label{12th equation}
    \mathrm{out}_i = \mathrm{add}(\mathrm{ReLU}(\mathrm{Conv}(\mathrm{Ref}_1)),\, M_i)
\end{equation}

\subsubsection{Attention Module}

\begin{figure}[h!]
    \centering
    \includegraphics[scale=0.4]{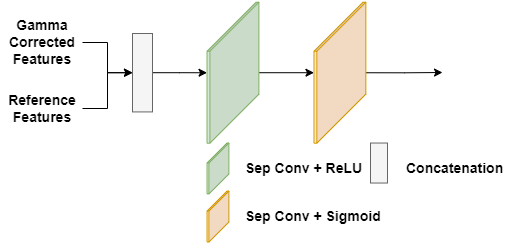}
    \caption{The structure of the Attention Module.}
    \label{fig: Fig6}
\end{figure}
The Attention Module is almost similar to \cite{9857192}, in which, as shown in Fig \ref{fig: Fig6}, feature maps are produced for Short- and Long-Exposure images to merge them with the reference image as guidance. After feeding the features of gamma-corrected images with the reference image, they are concatenated. Afterward, \(\mathrm{Sep}\mathrm{Conv+ReLU}\) and \(\mathrm{SepConv+Simgoid}\) operations are applied to them. The module can be considered as follows:
\begin{equation}\label{13th equation}
    R_i=\mathrm{ReLU}(\mathrm{SepConv}(\mathrm{concat}(f_i,f_r))\quad \textrm{for}\,\, i=1,3
\end{equation}
\begin{equation}\label{14th equation}
    S_i=\mathrm{Sigmoid}(\mathrm{SepConv}(R_i ))
\end{equation}
Where \(f_i\) and \(f_r\) are the features of gamma-corrected and reference images, respectively.

\subsubsection{Reconstruction}

\begin{figure}[!ht]
    \centering
    \includegraphics[scale=0.4]{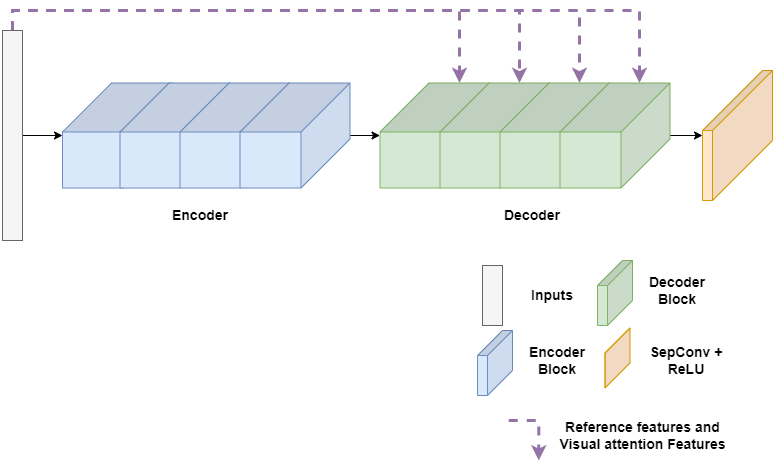}
    \caption{The total Scheme of the Reconstruction stage.}
    \label{fig: Fig7}
\end{figure}
All the extracted features from the modules are concatenated and fed to the reconstruction stage. As shown in Fig. \ref{fig: Fig7}, with the help of four encoder blocks, the input is merged, and new features are produced. Next, each decoder block receives features from the encoder along with features of the reference image and \ac{VAM}. Finally, a \(\mathrm{Sep\mathrm{Conv+ReLU}}\) is used to produce the output of the stage.

\begin{figure*}[!ht]
    \centering
    \includegraphics[scale=0.28]{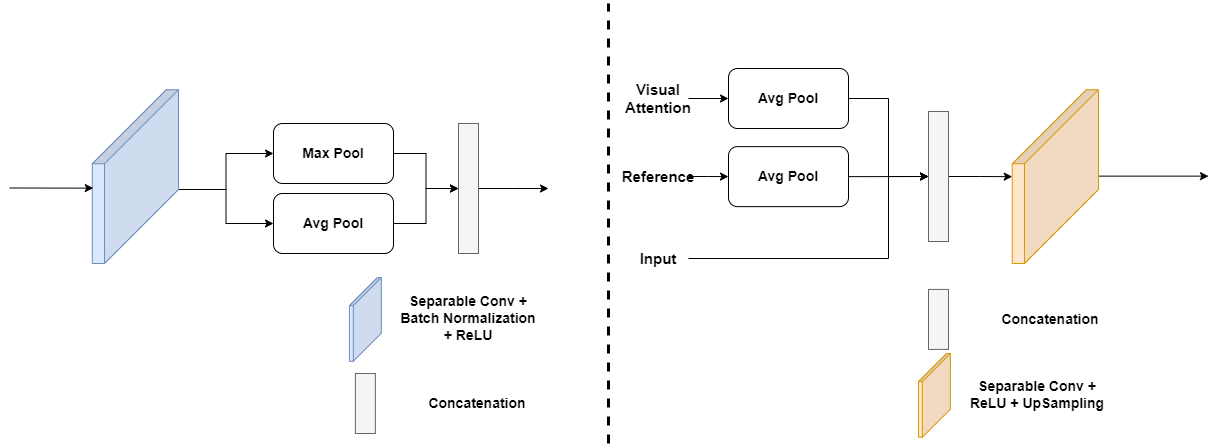}
    \caption{The structure of the blocks in the encoder (\textbf{left}) and the decoder (\textbf{right}).}
    \label{fig: Fig8}
\end{figure*}
Each encoder block (Fig. \ref{fig: Fig8}, left) initially applies \(\mathrm{SepConv}\), \(\mathrm{Batch\, Normalization}\), and \(\mathrm{ReLU}\) layers to the inputs. Afterward, similar to Feature Extraction Module, \(\mathrm{Max}\) and \(\mathrm{AVG}\) \(\mathrm{Poolings}\) are used. Finally, they are concatenated and sent to the next block.

Moreover, each decoder block (Fig. \ref{fig: Fig8}, right) consists of three inputs, which are features of the \ac{VAM}, features of the reference image, and the output of the previous block. First, \(\mathrm{AVG \,pooling}\) is applied to the first two inputs to make them the same size as the output of the previous block, and then they are concatenated with each other. Finally, \(\mathrm{Sep\mathrm{Conv+ReLU}}\) and \(\mathrm{Upsampling}\) are used, respectively.

\subsubsection{Refinement}

\begin{figure}[!ht]
    \centering
    \includegraphics[scale=0.35]{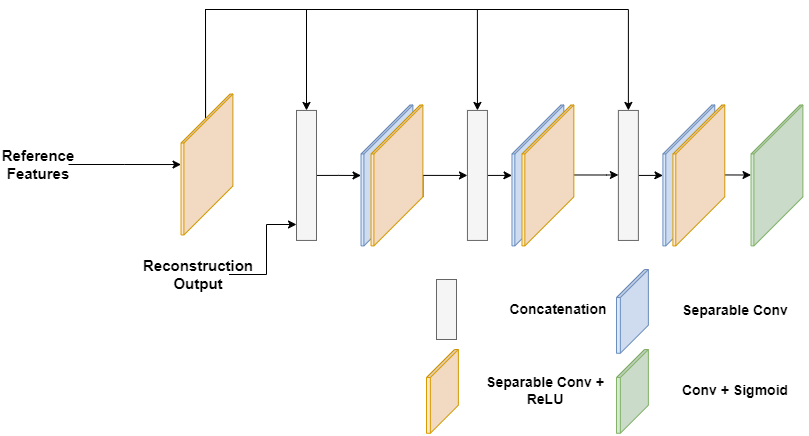}
    \caption{The structure of the Refinement Stage.}
    \label{fig: Fig9}
\end{figure}

Unfortunately, the output of the reconstruction stage may have blurry, saturated, or dark areas; therefore, to cope with such possible issues with the help of features of the reference image, a refinement section also has been added.

As Fig. \ref{fig: Fig9} illustrates, \(\mathrm{Sep\mathrm{Conv+ReLU}}\) is applied to the features of the reference image to reduce the number of feature maps. Furthermore, after concatenating the inputs, \(\mathrm{SepConv}\) and \(\mathrm{Sep\mathrm{Conv+ReLU}}\) are used, respectively. The process is repeated two more times, and eventually, \(Conv+Sigmoid\) is applied to produce the final image in Sigmoid space. The process in Refinement can be represented in pseudo-code as shown in Algorithm \ref{alg:Pseudo-code}.

\begin{algorithm}
\caption{Pseudo-code of Refinement Stage.}\label{alg:Pseudo-code}
\begin{algorithmic}
\State $\hat{f}_r=\mathrm{ReLU}(\mathrm{SepConv}(f_r))$
\State $i \gets 0$
\While{$i < 3$}
\If{$i == 0$}
    \State $c \gets concat(Reconstruction_o,\hat{f}_r)$
    \State $x \gets \mathrm{ReLU}(ConvSep(ConvSep(c)))$ 
\Else
    \State $c \gets concat(x,\hat{f}_r)$
    \State $x \gets  \mathrm{ReLU}(ConvSep(ConvSep(c)))$
\EndIf
\State $i\gets i + 1$
\EndWhile
\State $out \gets Sigmoid(Conv(x))$
\end{algorithmic}
\end{algorithm}

Notice that, in this research, the Ground Truth images are mapped from \ac{HDR} Space into sigmoid space. Indeed,  based on our experiments, transforming the values into sigmoid helps the network converge more conveniently. The reason for changing the space is that the values in \ac{HDR} space are too large, and a model with a low number of parameters is not able to learn to produce an \ac{HDR} image correctly; therefore, by mapping them to sigmoid space, the proposed model outperforms the model in the HDR space.

\section{Experiments and Results}
\label{sec:exp}

\subsection{Dataset}

A new dataset was collected for \ac{HDR} Imaging Challenge \cite{9523076, Perez-Pellitero_2022_CVPR}. In this dataset, two types of pictures (Single-Exposure and Multi-Exposure images) were provided; however, Multi-Exposure images only were used in this research. More specifically, this dataset includes images from \cite{10.1117/12.2040003} that were generated as follows. First, \ac{HDR} images were produced natively by two Alexa Arri cameras with a mirror rig; then, their corresponding \ac{LDR} images were generated synthetically with noise sources. There are approximately 1500 pairs of \ac{HDR}/\ac{LDR} images in this dataset for the training set, 40 for the validation set, and 200 pictures for the test set with a resolution of 1900x1060. However, in this research, we randomly selected 200 images of the training set as a test set and trained the model with around 1300 pairs.

\subsection{Implementation Details}

\begin{table}
    \centering
      \begin{tabular}{ |c |c| }
        \hline 
        Dataset & NTIRE Challenge \\
        \hline  
        Optimizer & Adam Optimizer \\ 
        \hline 
        Initial LR & 0.001 with LR decay  \\
        \hline 

        \begin{subtable}{0.2\textwidth}
            \centering
            \begin{tabular}[t]{c}
                \\
                Batch Size \\
                Input Size \\
                Augmentation
            \end{tabular}
        \end{subtable} & \begin{subtable}{0.3\textwidth}
            \centering
            \begin{tabular}[t]{c|c}
                Train & Validation \\
                 16 & 2 \\
                 256x256 & 1920x1088 \\
                 True & False
            \end{tabular}
        \end{subtable}  \\
        \hline 
        Epoch & 100  \\
        \hline 
        Loss & MAE \\
        \hline 
      \end{tabular}
      \caption{\label{table_1} Brief highlights regarding the training and validation settings for the proposed method.}
\end{table}

The highlights of the model are demonstrated in Table \ref{table_1} briefly. Additionally, the weights of the model were initialized randomly and no pre-trained weights were used. Finally, the information regarding the proposed method will be discussed in the following subsections.

\subsubsection{Loss function}

The \ac{MAE} loss function is used to train the model. The difference is that the Ground Truth is first mapped to Sigmoid Domain, and eventually, \ac{MAE} is calculated in Sigmoid space between the Ground Truth and the output of the model.

\begin{equation}\label{15th equation}
    GT_n=sigmoid(GT)	
\end{equation}
\begin{equation}\label{16th equation}
    L(\hat{y},GT_n )= |GT_n-\hat{y}|
\end{equation}
Where \(GT_n\) is the Ground Truth image in the new domain, and \(L\) is the loss between Ground Truth and the output.

Furthermore, after training the model in sigmoid space, inverse sigmoid is used to re-map the output to \ac{HDR} space. The inverse sigmoid can be written as follows:

\begin{equation}\label{17th:equation}
HDR=\log(\frac{\hat{(y)}}{1-\hat{y}})
\end{equation}

Where \(HDR\) is the output in \ac{HDR} space and \(\hat{y}\)  is the image in the sigmoid domain.

\subsubsection{Training}

Flipping the images vertically or horizontally is also used as an augmentation method during training. Moreover, before feeding the images to the model, they are resized into 256x256. The reason for doing so instead of producing patches is that some generated patches from the masks may be totally black or completely white, which causes the model to pay less attention to the images with Short-Exposure.

Moreover, batch size and the number of epochs are set to 16 and 100, respectively. In this article, Adam Optimizer with an initial learning of 0.001 is used, and it will be reduced by a factor of 0.1 if the validation accuracy does not improve. Finally, the whole model is implemented in Tensorflow (Keras) framework and is trained on a DGX-A100 GPU.

\subsubsection{Validation}
The images are first padded from 1900x1060 to 1920x1080 and then fed to the model without any augmentation methods during validation.

\subsection{Evaluation Metrics and Comparison}
\subsubsection{Quantitative Comparison}

The results in this paper are compared with the State-Of-The-Art methods by \(PSNR\)  in \ac{HDR} and Tone-mapped domains. The \(PSNR-\mu\) is the tone-mapped version, where the images were tone-mapped in \(\mu-law\). Moreover, the results are compared with the State-Of-The-Art methods in \(GMACs\) and the number of parameters.

\begin{table}
    \centering
      \begin{tabular}{ |c|c|c|c|c| }
        \hline 
        Methods & PSNR & \(\mu\)-PSNR & GMACs & Param. × $10^3$ \\
        \hline  
        GSANet & 36.88 & 35.57 &  \(\underline{199.38}\) &  \(\boldsymbol{80}\) \\ 
        \hline 
        DRHDR & 38.5 & \(\boldsymbol{36.91}\) & 1701.932 & 1190 \\
        \hline 
        Vien et al. & \(\underline{39.44}\) & 35.39 & \(\boldsymbol{198.819}\) & 1301 \\
        \hline 
        ours & \(\boldsymbol{43.25}\) & \(\underline{35.86}\) & 234.107 &  \(\underline{570}\)\\ 
        \hline 
      \end{tabular}
      \caption{\label{table_2} Comparison with the State-Of-The-Art methods. The bold numbers are the best values, and the underline ones are the second best.}
\end{table}

As mentioned in \cite{Perez-Pellitero_2022_CVPR}, the challenge focused on two tracks, which were Fidelity and low complexity. In the first one, the methods were required to obtain the highest \(\mu-PSNR\) while the \(GMACs\) value is less than 200. In the latter track, it was asked to reduce the \(GMACs\) value to less than the baseline method while the \(PSNR\) and \(\mu-PSNR\) values are almost the same as the baseline method. The proposed method has been compared with GSANet \cite{Li_2022_CVPR}, DRHDR \cite{MarinVega2022DRHDRAD}, and Vein et al. \cite{9856939} methods. As can be seen, Table \ref{table_2} shows the proposed method has the highest value in terms of \(PSNR\), while having the second highest value in \(\mu-PSNR\). On the other hand, Vien et al. \cite{9856939} had the lowest GMACs value, and GSANet is ranked second lowest. Moreover, it is visible that in terms of the number of parameters, GSANet has the lowest and the proposed method is in the second place among the algorithms.

\begin{table}[htbp]
    \centering
    \begin{tabular}{|c|c|c|}
    \hline
        Methods & PSNR & Mu-PSNR \\
    \hline
        Ours (HDR Space) & 42.4 & 35.28 \\
    \hline
        Ours (Sigmoid Space) & 43.25 & 35.86 \\
    \hline
    \end{tabular}
    \caption{\label{table_3}Comparison between the proposed method in \ac{HDR} and Sigmoid Spaces.}
\end{table}

Furthermore, for more study, the proposed method was trained and tested in \ac{HDR} and Sigmoid Spaces to check which space is superior for training the model. Thus, as Table \ref{table_3} demonstrates, the proposed method in Sigmoid Space outperformed the algorithm in the \ac{HDR} domain. Moreover, during training, the model in Sigmoid space converged quicker than the model in the \ac{HDR} domain.

\begin{figure*}[htbp]
    \centering
    \begin{subfigure}[b]{0.325\textwidth}
        \centering
        \includegraphics[width=\textwidth]{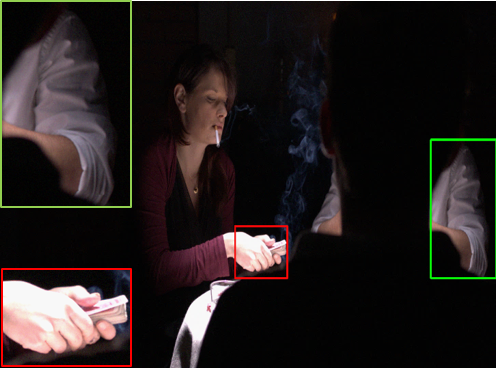}
    \end{subfigure}
    \begin{subfigure}[b]{0.325\textwidth}
        \centering
        \includegraphics[width=\textwidth]{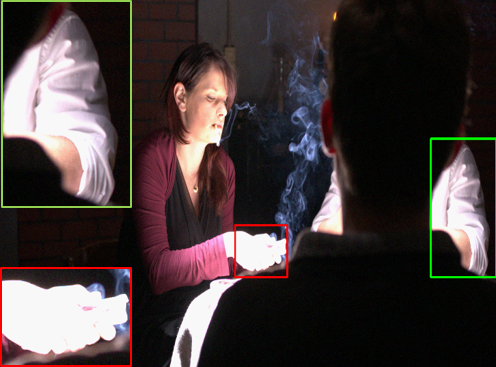}
    \end{subfigure}
    \begin{subfigure}[b]{0.325\textwidth}
        \centering
        \includegraphics[width=\textwidth]{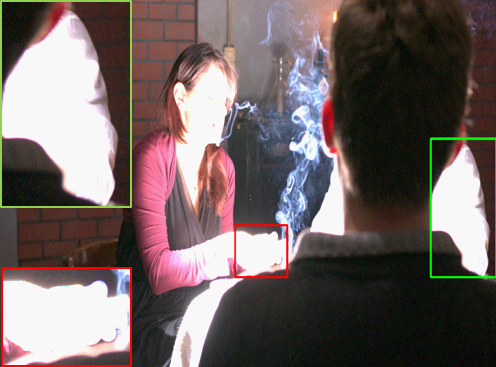}
    \end{subfigure}
    \begin{subfigure}[b]{0.242\textwidth}
        \centering
        \includegraphics[width=\textwidth]{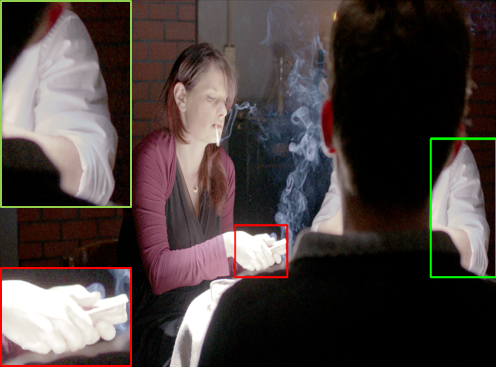}
    \end{subfigure}
    \begin{subfigure}[b]{0.242\textwidth}
        \centering
        \includegraphics[width=\textwidth]{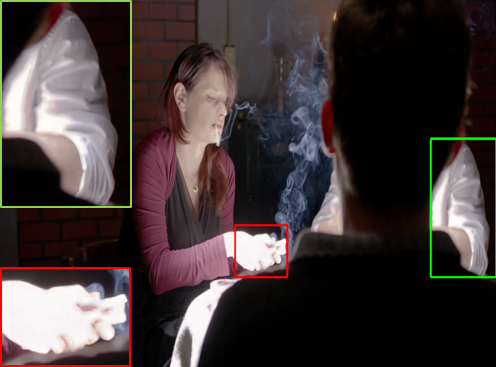}
    \end{subfigure}
    \begin{subfigure}[b]{0.242\textwidth}
        \centering
        \includegraphics[width=\textwidth]{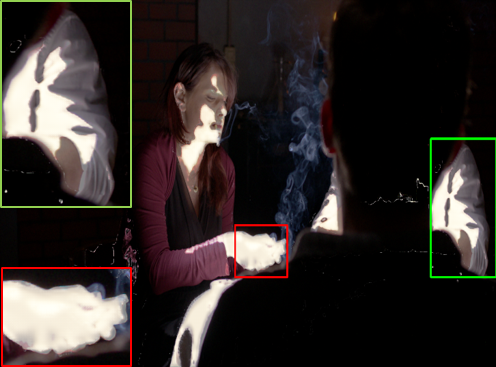}
    \end{subfigure}
    \begin{subfigure}[b]{0.242\textwidth}
        \centering
        \includegraphics[width=\textwidth]{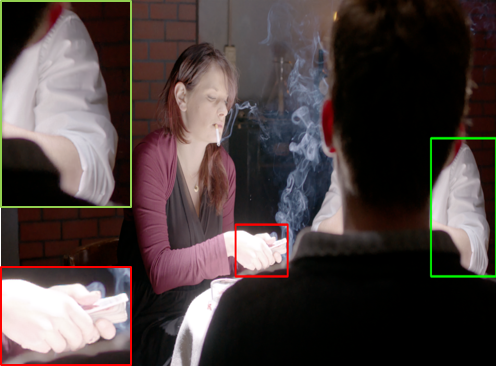}
    \end{subfigure}
    \begin{subfigure}[b]{0.325\textwidth}
        \centering
        \includegraphics[width=\textwidth]{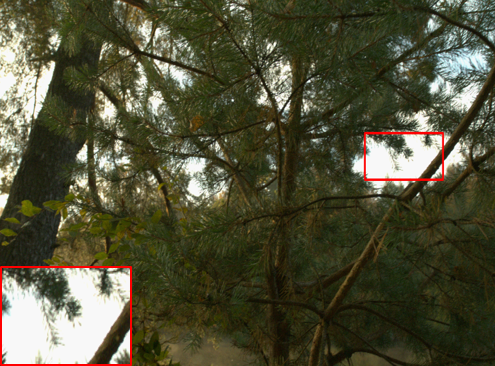}
    \end{subfigure}
    \begin{subfigure}[b]{0.325\textwidth}
        \centering
        \includegraphics[width=\textwidth]{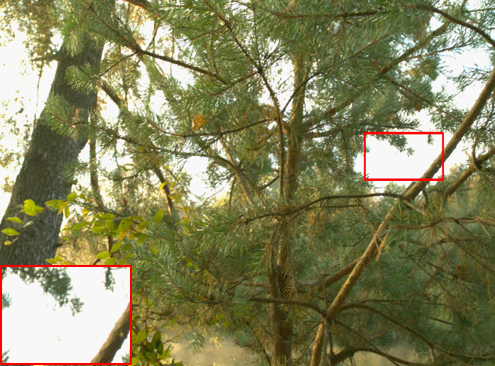}
    \end{subfigure}
    \begin{subfigure}[b]{0.325\textwidth}
        \centering
        \includegraphics[width=\textwidth]{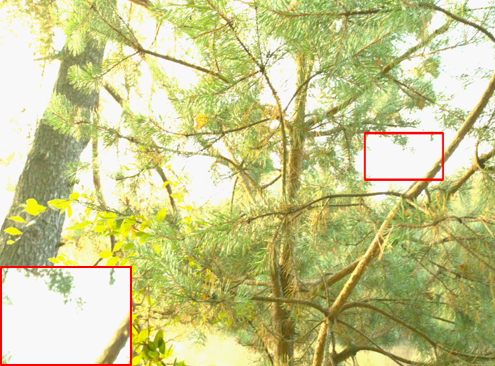}
    \end{subfigure}
    \begin{subfigure}[b]{0.242\textwidth}
        \centering
        \includegraphics[width=\textwidth]{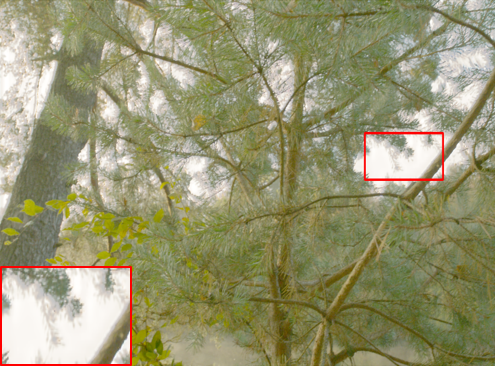}
    \end{subfigure}
    \begin{subfigure}[b]{0.242\textwidth}
        \centering
        \includegraphics[width=\textwidth]{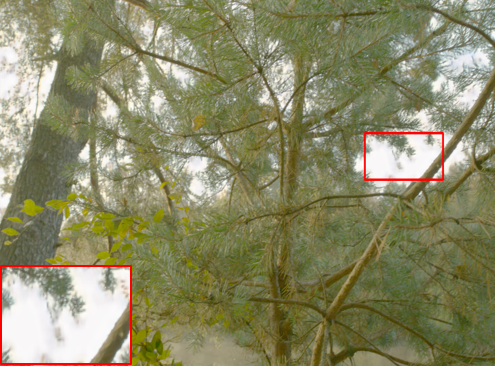}
    \end{subfigure}
    \begin{subfigure}[b]{0.242\textwidth}
        \centering
        \includegraphics[width=\textwidth]{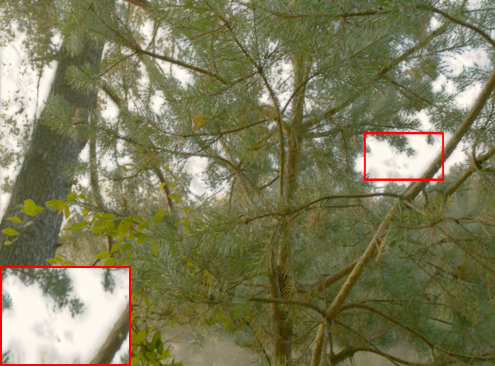}
    \end{subfigure}
    \begin{subfigure}[b]{0.242\textwidth}
        \centering
        \includegraphics[width=\textwidth]{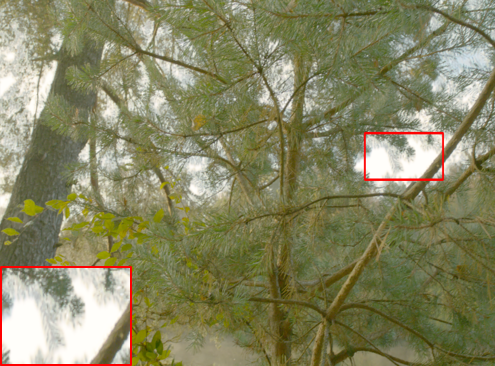}
    \end{subfigure}
    \begin{subfigure}[b]{0.325\textwidth}
        \centering
        \includegraphics[width=\textwidth]{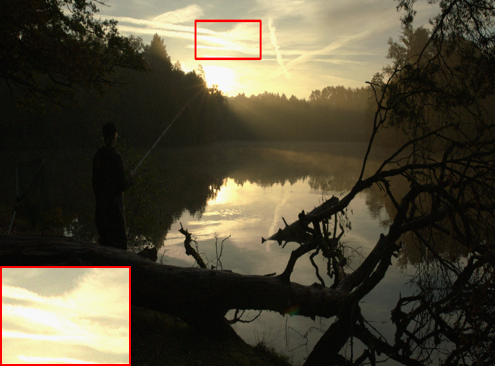}
    \end{subfigure}
    \begin{subfigure}[b]{0.325\textwidth}
        \centering
        \includegraphics[width=\textwidth]{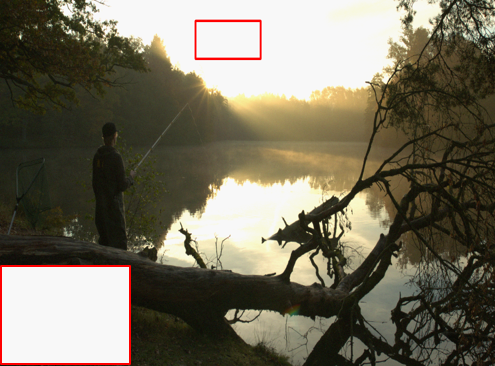}
    \end{subfigure}
    \begin{subfigure}[b]{0.325\textwidth}
        \centering
        \includegraphics[width=\textwidth]{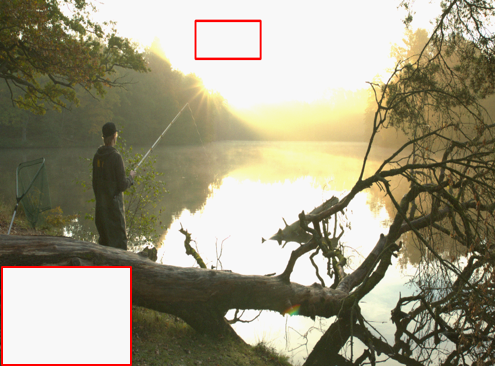}
    \end{subfigure}
    \begin{subfigure}[b]{0.242\textwidth}
        \centering
        \includegraphics[width=\textwidth]{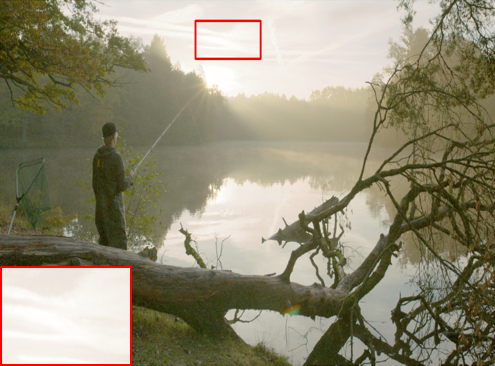}
    \end{subfigure}
    \begin{subfigure}[b]{0.242\textwidth}
        \centering
        \includegraphics[width=\textwidth]{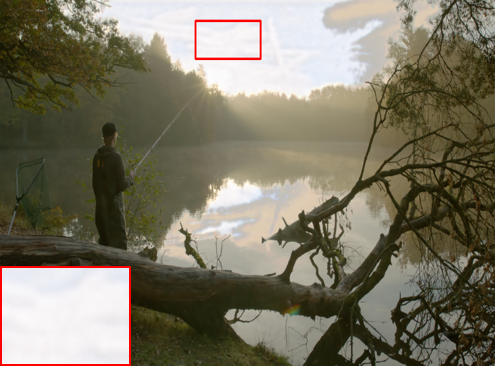}
    \end{subfigure}
    \begin{subfigure}[b]{0.242\textwidth}
        \centering
        \includegraphics[width=\textwidth]{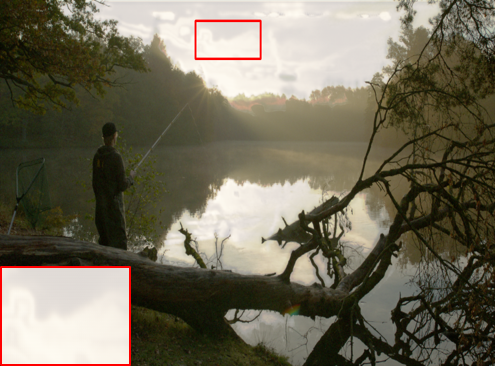}
    \end{subfigure}
        \begin{subfigure}[b]{0.242\textwidth}
        \centering
        \includegraphics[width=\textwidth]{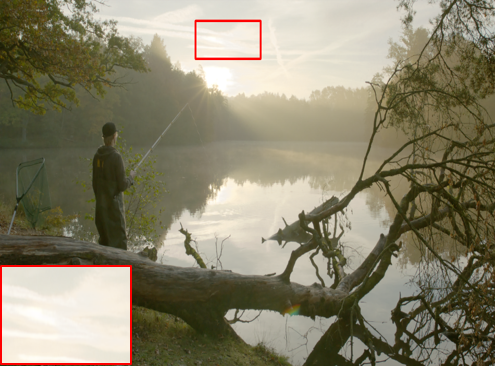}
    \end{subfigure}
    \caption{Qualitative Comparison with the State-Of-The-Art. The first row of each scene contains short, medium, and long exposure images, respectively. The second row includes ours, DRHDR, An et al., and GSANet outcomes, respectively.}
    \label{fig:Fig10}
\end{figure*}

\subsubsection{Qualitative Comparison}
As can be seen in Fig. \ref{fig:Fig10}, the produced images by ours, worked better in terms of image reconstruction compared to DRHDR and An et al. methods. More specifically, Fig. \ref{fig:Fig10} demonstrates the results of ours, DRHDR \cite{MarinVega2022DRHDRAD}, An et al. \cite{9856939}, and GSANet \cite{Li_2022_CVPR}. As can be seen, the output of An et al. in the first scene has distortion in the bright areas, and it is visible that the algorithm cannot restore the details from these areas correctly. Furthermore, there is some degradation in the dark regions too. Moreover, although DRHDR worked great and reconstructed both areas, this method was not able to acquire the details in over-saturated areas. For instance, looking at the two red and green boxes, the model did not reconstruct the details of the hands and the shirt, while the proposed method produced more detail in these two regions. Moreover, produced image from the GSANet method shows significant details and is almost similar to ours. More precisely, although both methods could reconstruct the shirt nicely, the details of the hand in the GSANet are more than ours.

Additionally, in the second scene, the DRHDR and An et al. methods were not able to reconstruct the branches that were only visible in the short exposure image and restored only a part of them. In contrast, the proposed method and the GSANet worked almost well in this regard. Finally, looking at the last scene, it is visible that the proposed method outperformed the first two algorithms and reconstructed more details in both dark and bright areas, and the details of the sky show this point.

Furthermore, although the segmentation helped the model to produce better results, the method might encounter two possible issues. Firstly, due to plausible noise in input images, using segmentation for extracting visible areas may also acquire the noise, and the produced image might become noisy. Lastly, although spatial alignment and attention modules are used to avoid any possible ghosting problems, if the input images have a severe amount of movement, the output might also encounter a ghosting issue. Because the segmentation is applied to the Short- and Long-Exposure images and extracts their visible areas. Therefore, some parts of the images might not be aligned. Moreover, for future research, we would like to investigate possible methods to use segmentation and avoid any likely noise or misalignment.

\section{Conclusion}
\label{sec:conclusions}
In this article, we proposed a new method for \ac{HDR} imaging with the help of image segmentation. More specifically, we first applied the Otsu method on Short- and Long-Exposure images to acquire the areas with more details. Afterward, the input images along with the segmentation outputs were fed to the model to produce the \ac{HDR} image. The results show that the proposed method outperformed the State-Of-The-Art and generated more details. However, the proposed model is not free of issues, and in case of possible noise or misalignment in input images, the output might have a slight amount of noise or misalignment due to extracting areas of input images. Therefore, for future research, we would like to focus on investigating these two problems.

\bibliographystyle{ieeetr}
\bibliography{bibliography}

\begin{thebibliography}{10}

\bibitem{855857}
S.~Nayar and T.~Mitsunaga, ``High dynamic range imaging: spatially varying
  pixel exposures,'' in {\em Proceedings IEEE Conference on Computer Vision and
  Pattern Recognition. CVPR 2000 (Cat. No.PR00662)}, vol.~1, pp.~472--479
  vol.1, 2000.

\bibitem{1467255}
J.~Tumblin, A.~Agrawal, and R.~Raskar, ``Why i want a gradient camera,'' in
  {\em 2005 IEEE Computer Society Conference on Computer Vision and Pattern
  Recognition (CVPR'05)}, vol.~1, pp.~103--110 vol. 1, 2005.

\bibitem{4118491}
M.~McGuire, W.~Matusik, H.~Pfister, B.~Chen, J.~F. Hughes, and S.~K. Nayar,
  ``Optical splitting trees for high-precision monocular imaging,'' {\em IEEE
  Computer Graphics and Applications}, vol.~27, no.~2, pp.~32--42, 2007.

\bibitem{10.1145/2010324.1964936}
M.~D. Tocci, C.~Kiser, N.~Tocci, and P.~Sen, ``A versatile hdr video production
  system,'' {\em ACM Trans. Graph.}, vol.~30, jul 2011.

\bibitem{Hajisharif2015}
S.~Hajisharif, J.~Kronander, and J.~Unger, ``Adaptive dualiso hdr
  reconstruction,'' {\em EURASIP Journal on Image and Video Processing},
  vol.~2015, p.~41, Dec 2015.

\bibitem{7168378}
H.~Zhao, B.~Shi, C.~Fernandez-Cull, S.-K. Yeung, and R.~Raskar, ``Unbounded
  high dynamic range photography using a modulo camera,'' in {\em 2015 IEEE
  International Conference on Computational Photography (ICCP)}, pp.~1--10,
  2015.

\bibitem{10.5555/3058909.3058929}
A.~Serrano, F.~Heide, D.~Gutierrez, G.~Wetzstein, and B.~Masia, ``Convolutional
  sparse coding for high dynamic range imaging,'' in {\em Proceedings of the
  37th Annual Conference of the European Association for Computer Graphics}, EG
  '16, (Goslar, DEU), p.~153–163, Eurographics Association, 2016.

\bibitem{10.1145/3130800.3130816}
G.~Eilertsen, J.~Kronander, G.~Denes, R.~K. Mantiuk, and J.~Unger, ``Hdr image
  reconstruction from a single exposure using deep cnns,'' {\em ACM Trans.
  Graph.}, vol.~36, nov 2017.

\bibitem{SHE2023105947}
L.~She, M.~Ye, S.~Li, Y.~Zhao, C.~Zhu, and H.~Wang, ``Single-image hdr
  reconstruction by dual learning the camera imaging process,'' {\em
  Engineering Applications of Artificial Intelligence}, vol.~120, p.~105947,
  2023.

\bibitem{le2023singlehdr}
P.-H. Le, Q.~Le, R.~Nguyen, and B.-S. Hua, ``Single-image hdr reconstruction by
  multi-exposure generation,'' in {\em Proceedings of the IEEE/CVF Winter
  Conference on Applications of Computer Vision (WACV)}, January 2023.

\bibitem{10.1145/3550277}
G.~Cao, F.~Zhou, K.~Liu, A.~Wang, and L.~Fan, ``A decoupled kernel prediction
  network guided by soft mask for single image hdr reconstruction,'' {\em ACM
  Trans. Multimedia Comput. Commun. Appl.}, vol.~19, feb 2023.

\bibitem{9116898}
A.~Omrani, M.~R. Soheili, and M.~Kelarestaghi, ``High dynamic range image
  reconstruction using multi-exposure wavelet hdrcnn,'' in {\em 2020
  International Conference on Machine Vision and Image Processing (MVIP)},
  pp.~1--4, 2020.

\bibitem{8457442}
S.~Lee, G.~H. An, and S.-J. Kang, ``Deep chain hdri: Reconstructing a high
  dynamic range image from a single low dynamic range image,'' {\em IEEE
  Access}, vol.~6, pp.~49913--49924, 2018.

\bibitem{10.1145/3130800.3130834}
Y.~Endo, Y.~Kanamori, and J.~Mitani, ``Deep reverse tone mapping,'' {\em ACM
  Trans. Graph.}, vol.~36, nov 2017.

\bibitem{8803582}
G.~R. K.S., A.~Biswas, M.~S. Patel, and B.~H.~P. Prasad, ``Deep multi-stage
  learning for hdr with large object motions,'' in {\em 2019 IEEE International
  Conference on Image Processing (ICIP)}, pp.~4714--4718, 2019.

\bibitem{10.1145/3072959.3073609}
N.~K. Kalantari and R.~Ramamoorthi, ``Deep high dynamic range imaging of
  dynamic scenes,'' {\em ACM Trans. Graph.}, vol.~36, jul 2017.

\bibitem{Wu_2018_ECCV}
S.~Wu, J.~Xu, Y.-W. Tai, and C.-K. Tang, ``Deep high dynamic range imaging with
  large foreground motions,'' in {\em Proceedings of the European Conference on
  Computer Vision (ECCV)}, September 2018.

\bibitem{Yan_2019_CVPR}
Q.~Yan, D.~Gong, Q.~Shi, A.~v.~d. Hengel, C.~Shen, I.~Reid, and Y.~Zhang,
  ``Attention-guided network for ghost-free high dynamic range imaging,'' in
  {\em Proceedings of the IEEE/CVF Conference on Computer Vision and Pattern
  Recognition (CVPR)}, June 2019.

\bibitem{8747329}
K.~R. Prabhakar, R.~Arora, A.~Swaminathan, K.~P. Singh, and R.~V. Babu, ``A
  fast, scalable, and reliable deghosting method for extreme exposure fusion,''
  in {\em 2019 IEEE International Conference on Computational Photography
  (ICCP)}, pp.~1--8, 2019.

\bibitem{10.1007/978-3-030-58589-1_30}
K.~R. Prabhakar, S.~Agrawal, D.~K. Singh, B.~Ashwath, and R.~V. Babu, ``Towards
  practical and efficient high-resolution hdr deghosting with cnn,'' in {\em
  Computer Vision -- ECCV 2020} (A.~Vedaldi, H.~Bischof, T.~Brox, and J.-M.
  Frahm, eds.), (Cham), pp.~497--513, Springer International Publishing, 2020.

\bibitem{8282319}
V.~G. An and C.~Lee, ``Single-shot high dynamic range imaging via deep
  convolutional neural network,'' in {\em 2017 Asia-Pacific Signal and
  Information Processing Association Annual Summit and Conference (APSIPA
  ASC)}, pp.~1768--1772, 2017.

\bibitem{8658831}
Q.~Yan, D.~Gong, P.~Zhang, Q.~Shi, J.~Sun, I.~Reid, and Y.~Zhang, ``Multi-scale
  dense networks for deep high dynamic range imaging,'' in {\em 2019 IEEE
  Winter Conference on Applications of Computer Vision (WACV)}, pp.~41--50,
  2019.

\bibitem{otsu1979threshold}
N.~Otsu, ``A threshold selection method from gray-level histograms,'' {\em IEEE
  transactions on systems, man, and cybernetics}, vol.~9, no.~1, pp.~62--66,
  1979.

\bibitem{Li_2022_CVPR}
F.~Li, R.~Gang, C.~Li, J.~Li, S.~Ma, C.~Liu, and Y.~Cao, ``Gamma-enhanced
  spatial attention network for efficient high dynamic range imaging,'' in {\em
  Proceedings of the IEEE/CVF Conference on Computer Vision and Pattern
  Recognition (CVPR) Workshops}, pp.~1032--1040, June 2022.

\bibitem{Deng2022}
Y.~Deng, Q.~Liu, and T.~Ikenaga, ``Attention-guided network with inverse
  tone-mapping guided up-sampling for hdr imaging of dynamic scenes,'' {\em
  Multimedia Tools and Applications}, vol.~81, pp.~12925--12944, Apr 2022.

\bibitem{MarinVega2022DRHDRAD}
J.~Mar'in-Vega, M.~Sloth, P.~Schneider-Kamp, and R.~Rottger, ``Drhdr: A dual
  branch residual network for multi-bracket high dynamic range imaging,'' {\em
  2022 IEEE/CVF Conference on Computer Vision and Pattern Recognition Workshops
  (CVPRW)}, pp.~843--851, 2022.

\bibitem{ye2022learning}
Q.~Ye, M.~Suganuma, J.~Xiao, and T.~Okatani, ``Learning regularized multi-scale
  feature flow for high dynamic range imaging,'' 2022.

\bibitem{xiao2022multiscale}
J.~Xiao, Q.~Ye, T.~Liu, C.~Zhang, and K.-M. Lam, ``Multi-scale sampling and
  aggregation network for high dynamic range imaging,'' 2022.

\bibitem{9857164}
Q.~Yan, S.~Zhang, W.~Chen, Y.~Liu, Z.~Zhang, Y.~Zhang, J.~Q. Shi, and D.~Gong,
  ``A lightweight network for high dynamic range imaging,'' in {\em 2022
  IEEE/CVF Conference on Computer Vision and Pattern Recognition Workshops
  (CVPRW)}, pp.~823--831, 2022.

\bibitem{9857192}
G.~Yu, J.~Zhang, Z.~Ma, and H.~Wang, ``Efficient progressive high dynamic range
  image restoration via attention and alignment network,'' in {\em 2022
  IEEE/CVF Conference on Computer Vision and Pattern Recognition Workshops
  (CVPRW)}, pp.~1123--1130, 2022.

\bibitem{dai2022waveletbased}
T.~Dai, W.~Li, X.~Cao, J.~Liu, X.~Jia, A.~Leonardis, Y.~Yan, and S.~Yuan,
  ``Wavelet-based network for high dynamic range imaging,'' 2022.

\bibitem{prabhakar2022segmentation}
K.~R. Prabhakar, S.~Agrawal, and R.~V. Babu, ``Segmentation guided deep hdr
  deghosting,'' 2022.

\bibitem{9856939}
A.~G. Vien, S.~Park, T.~T.~N. Mai, G.~Kim, and C.~Lee, ``Bidirectional motion
  estimation with cyclic cost volume for high dynamic range imaging,'' in {\em
  2022 IEEE/CVF Conference on Computer Vision and Pattern Recognition Workshops
  (CVPRW)}, pp.~1182--1189, 2022.

\bibitem{Messikommer_2022_CVPR}
N.~Messikommer, S.~Georgoulis, D.~Gehrig, S.~Tulyakov, J.~Erbach,
  A.~Bochicchio, Y.~Li, and D.~Scaramuzza, ``Multi-bracket high dynamic range
  imaging with event cameras,'' in {\em Proceedings of the IEEE/CVF Conference
  on Computer Vision and Pattern Recognition (CVPR) Workshops}, pp.~547--557,
  June 2022.

\bibitem{10.1007/978-3-031-19800-7_20}
Z.~Liu, Y.~Wang, B.~Zeng, and S.~Liu, ``Ghost-free high dynamic range imaging
  with context-aware transformer,'' in {\em Computer Vision -- ECCV 2022}
  (S.~Avidan, G.~Brostow, M.~Ciss{\'e}, G.~M. Farinella, and T.~Hassner, eds.),
  (Cham), pp.~344--360, Springer Nature Switzerland, 2022.

\bibitem{10.1007/978-3-540-30125-7_5}
A.~Vadivel, M.~Mohan, S.~Sural, and A.~K. Majumdar, ``Segmentation using
  saturation thresholding and its application in content-based retrieval of
  images,'' in {\em Image Analysis and Recognition} (A.~Campilho and M.~Kamel,
  eds.), (Berlin, Heidelberg), pp.~33--40, Springer Berlin Heidelberg, 2004.

\bibitem{8672126}
Y.~Kinoshita and H.~Kiya, ``Scene segmentation-based luminance adjustment for
  multi-exposure image fusion,'' {\em IEEE Transactions on Image Processing},
  vol.~28, no.~8, pp.~4101--4116, 2019.

\bibitem{kinoshita_kiya_2018}
Y.~Kinoshita and H.~Kiya, ``Automatic exposure compensation using an image
  segmentation method for single-image-based multi-exposure fusion,'' {\em
  APSIPA Transactions on Signal and Information Processing}, vol.~7, p.~e22,
  2018.

\bibitem{9568865}
B.~D. Lee and M.~H. Sunwoo, ``Hdr image reconstruction using segmented image
  learning,'' {\em IEEE Access}, vol.~9, pp.~142729--142742, 2021.

\bibitem{mitiche2010variational}
A.~Mitiche and I.~B. Ayed, {\em Variational and level set methods in image
  segmentation}, vol.~5.
\newblock Springer Science \& Business Media, 2010.

\bibitem{yi2012image}
F.~Yi and I.~Moon, ``Image segmentation: A survey of graph-cut methods,'' in
  {\em 2012 international conference on systems and informatics (ICSAI2012)},
  pp.~1936--1941, IEEE, 2012.

\bibitem{minaee2021image}
S.~Minaee, Y.~Boykov, F.~Porikli, A.~Plaza, N.~Kehtarnavaz, and D.~Terzopoulos,
  ``Image segmentation using deep learning: A survey,'' {\em IEEE transactions
  on pattern analysis and machine intelligence}, vol.~44, no.~7,
  pp.~3523--3542, 2021.

\bibitem{9594668}
L.~Wang and K.-J. Yoon, ``Deep learning for hdr imaging: State-of-the-art and
  future trends,'' {\em IEEE Transactions on Pattern Analysis and Machine
  Intelligence}, vol.~44, no.~12, pp.~8874--8895, 2022.

\bibitem{9523076}
E.~Pérez-Pellitero, S.~Catley-Chandar, A.~Leonardis, R.~Timofte, X.~Wang,
  Y.~Li, T.~Wang, F.~Song, Z.~Liu, W.~Lin, X.~Li, Q.~Rao, T.~Jiang, M.~Han,
  H.~Fan, J.~Sun, S.~Liu, , {\em et~al.}, ``Ntire 2021 challenge on high
  dynamic range imaging: Dataset, methods and results,'' in {\em 2021 IEEE/CVF
  Conference on Computer Vision and Pattern Recognition Workshops (CVPRW)},
  pp.~691--700, 2021.

\bibitem{Perez-Pellitero_2022_CVPR}
E.~P{\'e}rez-Pellitero, S.~Catley-Chandar, R.~Shaw, A.~Leonardis, R.~Timofte,
  Z.~Zhang, C.~Liu, Y.~Peng, Y.~Lin, G.~Yu, {\em et~al.}, ``Ntire 2022
  challenge on high dynamic range imaging: Methods and results,'' in {\em
  Proceedings of the IEEE/CVF Conference on Computer Vision and Pattern
  Recognition}, pp.~1009--1023, 2022.

\bibitem{10.1117/12.2040003}
J.~Froehlich, S.~Grandinetti, B.~Eberhardt, S.~Walter, A.~Schilling, and
  H.~Brendel, ``{Creating cinematic wide gamut HDR-video for the evaluation of
  tone mapping operators and HDR-displays},'' in {\em Digital Photography X}
  (N.~Sampat, R.~Tezaur, S.~Battiato, and B.~A. Fowler, eds.), vol.~9023,
  p.~90230X, International Society for Optics and Photonics, SPIE, 2014.

\end{thebibliography}

\end{document}